\pdfoutput=1
\documentclass[11pt]{article}

\usepackage[dvipsnames]{xcolor}

\usepackage{acl}

\usepackage{times}
\usepackage{latexsym}

\usepackage[T1]{fontenc}
\usepackage[utf8]{inputenc}

\usepackage{microtype}

\usepackage{multirow}
\usepackage{booktabs}
\usepackage{graphicx}
\usepackage{latexsym}
\usepackage{caption}
\usepackage{subcaption}
\usepackage{color}
\usepackage{soul}
\usepackage{graphicx}
\usepackage{svg}
\usepackage{textcomp}
\usepackage{url}
\usepackage{siunitx}
\usepackage{gensymb}
\usepackage{amsmath}
\usepackage{amssymb}% http://ctan.org/pkg/amssymb
\usepackage{pifont}% http://ctan.org/pkg/pifont

\definecolor{MyCyan}{HTML}{95FFF6}
\DeclareRobustCommand{\hlcyan}[1]{{\sethlcolor{MyCyan}\hl{#1}}}

\usepackage[normalem]{ulem} 
\usepackage{algorithm}
\usepackage{algpseudocode}
\makeatletter
\def\BState{\State\hskip-\ALG@thistlm}
\makeatother

\title{Context-Aware Document Simplification}

\author{Liam Cripwell \\ Université de Lorraine \\CNRS/LORIA \\ liam.cripwell@loria.fr \\  
\And Joël Legrand \\ Université de Lorraine \\ Centrale Supélec \\ CNRS/LORIA \\ joel.legrand@inria.fr\\
\And Claire Gardent \\ CNRS/LORIA  \\ Université de Lorraine \\ claire.gardent@loria.fr}

\begin{document}
\maketitle
\begin{abstract}
To date, most work on text simplification has focused on sentence-level inputs. Early attempts at document simplification merely applied these approaches iteratively over the sentences of a document. However, this fails to coherently preserve the discourse structure, leading to suboptimal output quality. Recently, strategies from controllable simplification have been leveraged to achieve state-of-the-art results on document simplification by first generating a document-level plan (a sequence of sentence-level simplification operations) and using this plan to guide sentence-level simplification downstream. However, this is still limited in that the simplification model has no direct access to the local inter-sentence document context, likely having a negative impact on surface realisation. We explore various systems that use document context within the simplification process itself, either by iterating over larger text units or by extending the system architecture to attend over a high-level representation of document context. In doing so, we achieve state-of-the-art performance on the document simplification task, even when not relying on plan-guidance. Further, we investigate the performance and efficiency tradeoffs of system variants and make suggestions of when each should be preferred.
\end{abstract}

\section{Introduction}
Text simplification transforms a given text into a simpler version of itself that can be understood by a wider audience, while preserving the same core meaning \citep{gooding-2022-ethical}. It has also proven useful as a preprocessing step for downstream NLP tasks such as machine translation \citep{chandrasekar-etal-1996-motivations,mishra-etal-2014-exploring,li-nenkova-2015-detecting,stajner-popovic-2016-text} and relation extraction \citep{miwa-etal-2010-entity,niklaus-etal-2016-sentence}.

Most previous work has focused on sentence-level simplification by training neural models on complex/simple sentence pairs under the assumption that they will learn to perform required operations (e.g. sentence splitting, lexical substitution or syntactic rephrasing) implicitly from the training data~\citep{zhang-lapata-2017-sentence,nisioi-etal-2017-exploring,jiang-etal-2020-neural}. However, the imbalanced representation of simplification operations throughout popular datasets, and the overly-conservative models arising from their use, have led to attempts at controllable simplification to achieve more variation and diversity in output texts~\cite{alva-manchego-etal-2017-learning,cripwell-etal-2021-discourse-based,maddela-etal-2021-controllable}.

Recently, strategies from controllable simplification have been leveraged to achieve state-of-the-art results on the document simplification task~\citep{cripwell-etal-2023-document}. Specifically, by using a planning model capable of considering the sentences surrounding a complex sentence, a sentence-level simplification model can be guided such that the structure of the resulting document remains more coherent. Despite this success, the sentence simplification model still has no direct access to document context which we believe limits the extent to which it can accurately produce simplified sentences that are consistent with the larger document. 

As such, we propose various systems that allow access to some representation of surrounding content within the simplification module, while still allowing for the possibility of plan-guidance. We show that in doing so, we are able to achieve state-of-the-art document simplification performance on the Newsela dataset, even without relying on a generated plan. Further, we investigate the performance and efficiency tradeoffs of various system variants.~\footnote{Pretrained models, code, and data are available at \url{https://github.com/liamcripwell/plan_simp}.}

Our key contributions are (i) a detailed investigation of how document context, input text and simplification plans impact document-level simplification and (ii) several state of the art models for document simplification. We show in particular that document level simplification is improved by combining a representation of the local context surrounding complex sentences with a simplification plan indicating how complex sentences should be simplified (whether they should be deleted, rephrased, split or copied).

\section{Related Work}

\paragraph{Context in Controlled Text Generation}
The use of external context within controlled text generation pipelines has seen recent success in areas outside of simplification. \citet{junyi_kbreview} control review generation by using document and sentence-level plans in the form of knowledge graph subgraphs. \citet{smith2020controlling} control the style of generated dialog responses by conditioning on a desired style token appended to other contextual utterances. \citet{hazarika2022zeroshot} modulate the amount of attention paid to different parts of a dialog context and show that using contextual encoding of question phrases can guide a model to more often generate responses in the form of questions. \citet{slobodkin2022controlled} consider summarisation where salient spans are first identified before being used to control the generation, while \citet{narayan2023conditional} first generate a summarisation plan consisting of question-answer pairs.

\paragraph{Simplification Planning}
Certain controllable sentence simplification works have approached simplification as a planning problem whereby an operation plan is first generated before being realised downstream to form the simplified text. The first of these are revision-based models that predict a sequence of token-level operations (delete, substitute, etc.), allowing for more control and interpretability \cite{alva-manchego-etal-2017-learning,dong-etal-2019-editnts,kumar-etal-2020-iterative,omelianchuk-etal-2021-text,dehghan-etal-2022-grs}. Others have taken a sentence-level approach by predicting a high-level operation (sentence split, rephrase, etc.) and using this to condition more typical neural systems \cite{scarton-specia-2018-learning,scarton2020,garbacea-etal-2021-explainable,cripwell-etal-2022-controllable}.

Recently, the latter approach was leveraged for document simplification where it obtained state-of-the-art performance~\cite{cripwell-etal-2023-document}. Here, a sequence of sentence-level operations is predicted for an entire document and then used to iteratively condition a sentence-level simplification model. The system considers both local (token representation of the sentence) and global document context (sequence of sentence-level encodings) when predicting an operation for a given sentence.

\paragraph{Document-Level Simplification.}
Initial attempts at document simplification simply applied sentence simplification methods iteratively over documents \citep{Woodsend2011WikiSimpleAS,alva-manchego-etal-2019-cross,sun-etal-2021-document}. However, it was noted this alone is insufficient for performing certain operations, often leading to poor discourse coherence in the output \citep{siddharthan-2003-preserving,alva-manchego-etal-2019-cross}.

Various sub-problems of document simplification have been approached in isolation, such as sentence deletion \citep{Zhong_Jiang_Xu_Li_2020,zhang-etal-2022-predicting}, insertion \citep{srikanth-li-2021-elaborative}, and re-ordering \citep{lin-etal-2021-towards-document-level}. \citet{sun-etal-2021-document} took a holistic approach by iteratively applying a sentence-level model, but with additional encoders to embed the two preceding and following sentences, which are used as additional input during generation. However, this was unable to outperform baselines.

Recently, \citet{cripwell-etal-2023-document} achieved state-of-the-art performance by producing a document simplification framework capable of performing all of the most common operations. Specifically, they use both high-level document context and sentence-level features to generate a plan specifying which operations to be performed on each sentence in a given document, which is then used to condition a sentence simplification model.

\section{Problem Formulation}

The goal of text simplification is to generate a text $S$ that simplifies
an input text $C$. In the document-level case, $C = c_1 \ldots c_n$ is a sequence of complex sentences and $S = s_1 \ldots s_m$ is a sequence of simple sentences. \citet{cripwell-etal-2023-document} further decompose this task into a two-stage process wherein a generated plan conditions the simplification:

$$P(S \mid C) = P(S \mid C, O) P(O \mid C)$$

where $O = o_1 \ldots o_n$ is a simplification plan, i.e. a sequence of sentence-level simplification operations for $C$ (\textit{copy}, \textit{rephrase}, \textit{split}, or \textit{delete}). The motivation here is that the plan provides a high-level description of how to transform $C$ into $S$, which can in turn be used to guide the iterative generation of the simplified document across sentences.

%CG
Although the use of such plans has shown improved results, little attention has been given to how the generation stage itself can be modified to improve document-level simplification. In this work, we investigate whether further changes can be made to simplification models in order to make better use of high-level plans, or alternatively, whether it is possible to forego the planning stage entirely by incorporating high-level document context into the generative model directly.

\paragraph{Terminology and Notations.} We use the following terminology and notational conventions:
\begin{itemize}
  \item $C = c_1 \ldots c_n$ is a complex document of $n$ sentences;
  \item $p_i$ is the $i$th paragraph from the complex document C;
  \item $S = s_1 \ldots s_m$ is a ground-truth simplified version of $C$, containing $m$ sentences;
  \item $\hat{S} = \hat{s}_1 \ldots \hat{s}_{m'}$ is a predicted simplification of $C$, generated by a simplification model;
  \item $o$ is a simplification operation with value \textit{copy}, \textit{rephrase}, \textit{split}, or \textit{delete};
  \item $\hat{O} = \hat{o}_1 \ldots \hat{o}_n$ is a predicted simplification plan stipulating specific sentence-level operations that should be applied to each $c_i \in C$ so as to arrive at some $\hat{S}$;
  \item $Z_i$ is a high-level representation of the document context for $c_i$. It is a sequence of vector encodings for a fixed window of sentences surrounding $c_i$ within $C$.
\end{itemize}

\section{Data}
For all experiments, we use Newsela-auto \cite{jiang-etal-2020-neural} which is currently the highest-quality document-level simplification dataset available. It consists of 1,130 English news articles from the original Newsela~\cite{xu-etal-2015-problems} dataset which are each manually rewritten at five different levels of simplification, corresponding to discrete reading levels (0-4) of increasing simplicity. It also includes both sentence and paragraph alignments for each document pair. Like previous works, for all our models we prepend a control-token to the input specifying the target document reading level.

We use the same filtered version of Newsela-auto used in \citet{cripwell-etal-2023-document}, along with the same train/validation/test splits to allow for model comparison. This also includes plan labels, consisting of an operation (\textit{copy}, \textit{rephrase}, \textit{split}, or \textit{delete}) assigned to each sentence pair. Statistics of this data can be seen in Table~\ref{tab:data_stats}.

\begin{table}[!htbp]
  \centering
  \small
  \begin{tabular}{c|ccc}
  \toprule
   & Train & Validation & Test \\
  \midrule
  \# Document Pairs & 16,946 & 457 & 916 \\
  \# Paragraph Pairs & 335,018 & 9,061 & 17,885 \\
  \# Sentence Pairs & 654,796 & 17,688 & 35,292 \\
  \bottomrule
  \end{tabular}
  \caption{\textbf{Statistics of the filtered Newsela-auto dataset from \citet{cripwell-etal-2023-document}}. There is a train/validation/test split of 92.5\%/2.5\%/5\%, assigned at the document-level (i.e. sentences and paragraphs from the same document will be contained within the same set). All reading-level variations of a specific article are also contained within the same set. }
  \label{tab:data_stats}
\end{table}

\section{Models}

We distinguish three model categories: (i) models whose sole input is text and which simplify a document either by iterating over its sentences/paragraphs or by handling the entire document as a single input; (ii) models that take both a complex sentence and some representation of its document context as input and simplify a document by iterating over its sentences; and (iii) models that are guided by a plan via control-tokens denoting sentence-level simplification operations prepended to the input sequence. These are illustrated in Table~\ref{tab:model_inputs} and presented in more detail in the following subsections. Additional training details are outlined in Appendix~\ref{app:train_details}.

%We distinguish three model categories based on their inputs: (i) those that only take a text sequence \cg{from the input document as input and simplify a document by iterating over its text sequences (sentences or paragraph) or handling the whole document in one go }; (ii) those that take \sout{ a text sequence} \cg{a complex sentence and some representation of its document context and simplify a document by iterating over its sentences}; and (iii) those that are also guided by a plan via control-tokens prepended to the input sequence. These are illustrated in Table~\ref{tab:model_inputs}. Additional training details are outlined in Appendix~\ref{app:train_details}.

\begin{table*}[!htbp]
  \centering
  \small
  \begin{tabular}{lccccccc}
  \toprule
  \bf System & \multicolumn{7}{c}{\bf Input} \\
  \midrule
   & \multicolumn{3}{c}{Text} & Context & \multicolumn{3}{c}{Plan} \\
  \midrule
   & \it Document & \it Paragraph & \it Sentence &  & \it Document & \it Paragraph & \it Sentence \\
  \midrule
  BART & $C$ & $p_i$ & $c_i$ & - & - & - & - \\
  LED & $C$ & $p_i$ & - & - & - & - & - \\
  \midrule
  ConBART & - & - & $c_i$ & $Z_i$ & - \\
  \midrule
  $\text{PG}_{\text{Dyn}}$ (\citeyear{cripwell-etal-2023-document}) & - & - & $c_i$ & - & - & - & $\hat{o}_i$ \\
  $\hat{O} \rightarrow \text{BART}$ & $C$ & $p_i$ & $c_i$ & - & $\hat{O}$ & $\hat{o}_{j..j+|p_i|}$ & $\hat{o}_i$ \\
  $\hat{O} \rightarrow \text{LED}$ & $C$ & $p_i$ & - & - & $\hat{O}$ & $\hat{o}_{j..j+|p_i|}$ & - \\
  $\hat{O} \rightarrow \text{ConBART}$ & - & - & $c_i$ & $Z_i$ & - & - & $\hat{o}_i$ \\
  \bottomrule
  \end{tabular}
  \caption{\textbf{Different system types} and the specific forms of text, context, and plan inputs they consume. $C$ is a complex document, $c_i$ is the $i$th sentence of $C$, and $p_i$ is the $i$th paragraph of $C$. $\hat{O}$ is a predicted document simplification plan, $\hat{o}_i$ is the individual operation predicted for the $i$th sentence, and $\hat{o}_{j..j+|p_i|}$ is the plan extract for a specific paragraph $p_i$, where $j$ is the index of the first sentence in $p_i$.}
  \label{tab:model_inputs}
\end{table*}

\subsection{Text-Only Models}
The most basic group of models we test are those that simply take a text sequence as input. We use baseline models trained to take entire documents or individual sentences. We also experiment with using paragraph inputs, the results of which we believe should scale better to the document-level than isolated sentences. Because paragraphs contain a wider token-level representation of local context this might provide enough information to maintain coherency in the discourse structure of the final document.

\paragraph{BART.} We finetune BART~\cite{lewis-etal-2020-bart} to perform simplification at the document (\textbf{$\text{BART}_{\text{doc}}$}), sentence (\textbf{$\text{BART}_{\text{sent}}$}), and paragraph (\textbf{$\text{BART}_{\text{para}}$}) levels.~\footnote{All models are initialised with the pretrained \textit{facebook/bart-base} model from \url{https://huggingface.co/facebook/bart-base}.} Both $\text{BART}_{\text{sent}}$ and $\text{BART}_{\text{para}}$ are applied iteratively over a document and outputs are concatenated to form the final simplification result.

\paragraph{Longformer.} Encoder-decoder models like BART often produce worse outputs and become much slower the longer the input documents are. Longformer~\citep{longformer} is one proposal that aims to overcome these limitations by using a modified self-attention mechanism that scales linearly with sequence length. We finetune a Longformer encoder-decoder to perform the simplification on documents (\textbf{LED$_{\text{doc}}$}) and paragraphs (\textbf{LED$_{\text{para}}$}).~\footnote{All models are initialised with the pretrained \textit{allenai/led-base-16384} model from \url{https://huggingface.co/allenai/led-base-16384}.}

\subsection{Context-Aware Model (ConBART)}
\begin{figure}[!htbp]
  \centering
  \includegraphics[width=0.45\textwidth,angle=270,origin=c]{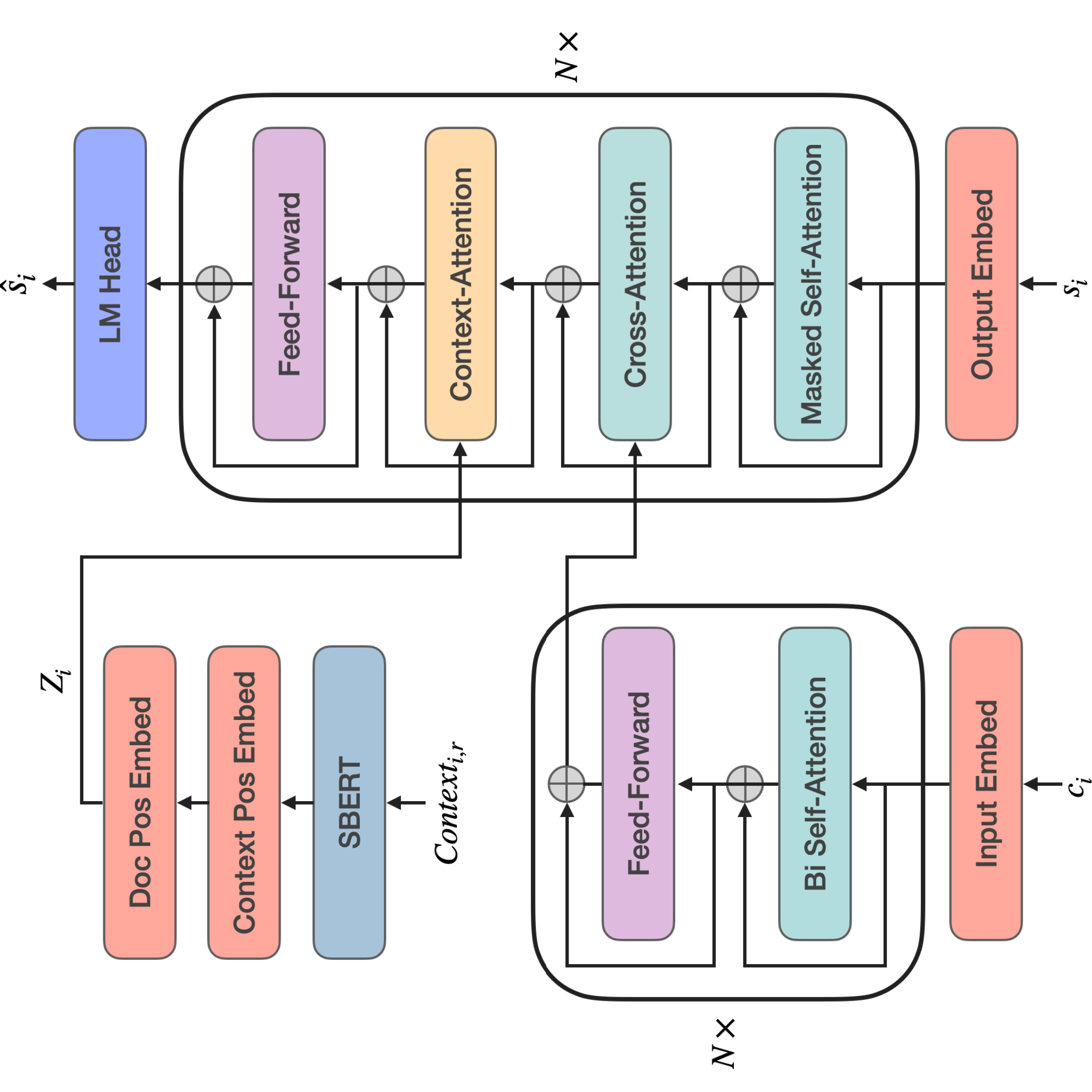}
  \caption{\textbf{ConBART model architecture.} The added context attention layer is shown in yellow, which allows for cross-attention over high-level document content, $Z_i$.}
  \label{fig:conbart_arch}
\end{figure}

We propose a context-aware modification of the BART architecture (\textbf{ConBART}) that is able to condition its generation on both an input sentence $c_i$ and a high-level representation of its document context $Z_i$ (a sequence of vectors representing surrounding sentences in the document). This is done via extra cross-attention layers in each decoder attention block that specifically focus on $Z_i$. The ConBART architecture is illustrated in Figure~\ref{fig:conbart_arch}.

We produce $Z_i$ by employing the same context representation strategy used for planning in \citet{cripwell-etal-2023-document}. Specifically, the document context is obtained by taking a fixed window of sentences surrounding the target $c_i$, encoding them with Sentence-BERT (SBERT, ~\citep{reimers-gurevych-2019-sentence}), and applying custom positional embeddings to represent location within the document. 

By generating the plan autoregressively, it is also possible to use previously simplified sentences within the left context of the current complex sentence, a method we refer to as \textit{dynamic context}. In this case, the window of sentences represented within $Z_i$ is defined:
\begin{equation}
  \label{eq:dyn_context}
  \begin{split}
  Context_{i,r} = & \mathit{Concat}(\hat{s}_{i-r..i-1}, c_{i..i+r})
  \end{split}
\end{equation}
where $r$ is the context window radius and $\hat{s}_i$ is the simplification output for the $i$th sentence $c_i$. We use the same recommended setting of $r=13$. 

The intuition behind the ConBART architecture is that the contextual information should allow for the simplification model to implicitly learn useful features of the discourse structure of the document in a similar way to the planner in \citet{cripwell-etal-2023-document}. 
%\vspace{-0.7cm}

\subsection{Plan-Guided Systems}
\begin{table*}[!htbp]
    \centering
    \small
    \begin{tabular}{llll|lll|cl|ll}
    \toprule
    \bf System & \multicolumn{3}{c}{\bf BARTScore $\uparrow$} & \multicolumn{3}{c}{\bf SMART $\uparrow$} & \bf FKGL $\downarrow$ & \bf SARI $\uparrow$  & \multicolumn{2}{c}{\bf Length} \\
    \midrule
      & P & R & F1 & P & R & F1 &  &  & Tok. & Sent. \\
      & ($r \rightarrow h$) & ($h \rightarrow r$) & & & & & & & &  \\
    \midrule
    Input & -2.47 & -1.99 & -2.23 & 63.2 & 62.7 & 62.8 & 8.44 & 20.5 & 866.9 & 38.6 \\
    Reference & -0.93 & -0.93 & -0.93 & 100 & 100 & 100 & 4.93 & 99.9 & 671.5 & 42.6 \\
    \midrule
    $\text{BART}_{\text{doc}}$ & -2.68 & -2.76 & -2.72 & 61.9 & 43.9 & 50.6 & 10.01 & 47.1 & 600.8 & 20.7 \\
    $\text{BART}_{\text{sent}}$ & -1.63 & -1.56 & -1.60 & 78.9 & 80.1 & 79.3 & 5.03 & 73.0 & 666.4 & 42.6 \\
    $\text{BART}_{\text{para}}$ & -1.85 & -1.49$^{\ast}$ & -1.67 & 77.2 & 82.8$^{\ast}$ & 79.6 & 5.28 & 73.7 & 752.8 & 45.6$^{\ast}$ \\
    $\text{LED}_{\text{doc}}$ & -1.68 & -1.73 & -1.70 & 75.3 & 74.9 & 74.8 & 4.87 & 68.7 & 643.7 & 41.5 \\
    $\text{LED}_{\text{para}}$ & -1.61 & -1.40$^{\ast}$ & -1.50$^{\ast}$ & 81.1 & \bf85.5$^{\ast}$ & 83.0$^{\ast}$ & 5.15 & 76.9$^{\ast}$ & 712.9 & 44.9$^{\ast}$ \\
  \midrule
  $\text{ConBART}$ & -1.59 & -1.50 & -1.54$^{\ast}$ & 81.2 & 82.5$^{\ast}$ & 81.7 & 5.01 & 75.8 & 669.8 & 42.8 \\
    \midrule
    $\text{PG}_{\text{Dyn}}$ (\citeyear{cripwell-etal-2023-document}) & -1.60 & -1.54 & -1.57 & 80.2 & 81.0 & 80.5 & 4.98 & 75.0 & 667.2 & 42.6 \\
    $\hat{O} \rightarrow \text{ConBART}$ & -1.52$^{\ast}$ & -1.45$^{\ast}$ & -1.48$^{\ast}$ & 82.8$^{\ast}$ & 84.0$^{\ast}$ & 83.2$^{\ast}$ & 4.96 & 78.3$^{\ast}$ & 671.6 & 43.0 \\
    $\hat{O} \rightarrow \text{BART}_{\text{para}}$ & -1.75 & -1.47$^{\ast}$ & -1.61 & 79.4 & 81.9 & 80.4 & 5.11 & 74.9 & 715.3 & 42.7 \\
    $\hat{O} \rightarrow \text{LED}_{\text{para}}$ & \bf-1.50$^{\ast}$ & \bf-1.42$^{\ast}$ & \bf-1.46$^{\ast}$ & \bf83.7$^{\ast}$ & 84.9$^{\ast}$ & \bf84.1$^{\ast}$ & 5.09 & \bf78.5$^{\ast}$ & 683.1 & 42.8 \\
    \midrule
    $\text{PG}_{\text{Oracle}}$ (\citeyear{cripwell-etal-2023-document}) & -1.39$^{\ast}$ & -1.40$^{\ast}$ & -1.40$^{\ast}$ & 85.5$^{\ast}$ & 85.0$^{\ast}$ & 85.3$^{\ast}$ & 4.91 & 80.7$^{\ast}$ & 655.6 & 42.1 \\
    $O \rightarrow \text{ConBART}$ & \bf-1.32$^{\ast}$ & \bf-1.32$^{\ast}$ & \bf-1.32$^{\ast}$ & \bf88.0$^{\ast}$ & \bf87.7$^{\ast}$ & \bf87.8$^{\ast}$ & 4.92 & \bf83.8$^{\ast}$ & 659.6 & 42.3 \\
    $O \rightarrow \text{BART}_{\text{para}}$ & -1.60 & -1.36$^{\ast}$ & -1.48$^{\ast}$ & 83.6$^{\ast}$ & 85.3$^{\ast}$ & 84.3$^{\ast}$ & 5.07 & 79.7$^{\ast}$ & 706.2 & 42.3 \\
    $O \rightarrow \text{LED}_{\text{para}}$ & -1.36$^{\ast}$ & -1.33$^{\ast}$ & -1.35$^{\ast}$ & 87.0$^{\ast}$ & 87.3$^{\ast}$ & 87.1$^{\ast}$ & 5.03 & 82.3$^{\ast}$ & 673.6 & 42.4 \\
    \bottomrule
    \end{tabular}
    \caption{\textbf{Results of document simplification systems on Newsela-auto.} For BARTScore, $h$ is the hypothesis and $r$ is the reference. Scores significantly higher than $\text{PG}_{\text{Dyn}}$ are denoted with $\ast$ ($p < 0.005$). Significance was determined with Student's $t$-tests.}
    \label{tab:simp_eval}
\end{table*}
\paragraph{Existing System.} We compare with the state-of-the-art system proposed by \citet{cripwell-etal-2023-document}, \textbf{$\text{PG}_{\text{Dyn}}$}, which consists of a standard sentence-level BART model that is guided by a planner, which predicts the simplification operation to be applied each input sentence given a left and right context window of sentence representations, $Z_i$. The planner uses dynamic document context, allowing it to auto-regressively update the left context part of $Z_i$ during planning as each sentence is simplified (see Equation~\ref{eq:dyn_context}). 

\paragraph{Pipelines.} We construct pipeline systems that consist of each of our proposed models, guided by a document plan generated by the planner from \citet{cripwell-etal-2023-document} (the same as is used by $\text{PG}_{\text{Dyn}}$). For this, we use modified versions of each simplification model that are trained to take an operation control-token at the beginning of each text input.

We refer to each of these pipeline systems as \textbf{$\hat{O} \rightarrow h$}, where $h$ is the simplification model. We also report results where the ground-truth/oracle plans are used to condition models (\textbf{$O \rightarrow h$}).

Note that because the planner updates its document context autoregressively at the sentence-level, this does not interface perfectly with paragraph-level simplification models. As such, for pipelines using a paragraph-level simplification model ($\hat{O} \rightarrow \text{BART}_{\text{para}}$, $\hat{O} \rightarrow \text{LED}_{\text{para}}$), we only update the planner's context after each paragraph has been processed. Thus, for those paragraph level models, the left context of a complex sentence $c_i$ is only simplified up to the first sentence of the paragraph containing $c_i$, i.e.

%sentences not at the beginning of a paragraph will have their plan predicted with slightly imperfect leftward context that contains some complex sentences:
\begin{equation}
  \begin{split}
  Context_{i,r} = & \mathit{Concat}(\hat{s}_{i-r..j-1}, c_{j..i+r})
  \end{split}
\end{equation}
where $j$ is the index of the first sentence within the same paragraph as $c_i$, assuming $j > i-r$.

We also experimented with multi-task systems that are trained to perform both planning and simplification within a single model, therefore not requiring a pipeline setup. However, this ultimately proved unsuccessful (further details in Appendix~\ref{app:mtl}).

% We use the document simplification planner from \citet{cripwell-etal-2023-document}. This is a modified version of the RoBERTa architecture that includes a cross-attention layer so document context can be conditioned upon while predicting a simplification operation for each sentence in a document. Document context is represented as a sequence of SBERT embeddings. A context window with a radius of 13 sentences is used. The left-context is dynamically updated to include simplifications from previous timesteps (dynamic context).

\begin{table*}[!htbp]
  \centering
  \small
  \begin{tabular}{l|lll|lll|lll|l}
  \toprule
  \bf System & \multicolumn{3}{c}{\bf Fluency} & \multicolumn{3}{c}{\bf Adequacy} & \multicolumn{3}{c}{\bf Simplicity} & \bf Mean \\
  \midrule
   & Minor & Major & All & Minor & Major & All & Minor & Major & All \\
  \midrule
  Reference & 90.9$^{\ast}$ & \bf96.0 & 93.4 & 80.8 & 70.7$^{\ast}$ & 75.8 & 83.8$^{\ast}$ & 82.8$^{\ast}$ & 83.3 & 84.2 \\
  \midrule
  $\text{PG}_{\text{Dyn}}$ (\citeyear{cripwell-etal-2023-document}) & 91.9$^{\ast}$ & 94.9 & 93.4 & \bf83.8 & 73.7 & 78.8 & 88.9 & 85.9 & 87.4 & 86.5 \\
  $\text{LED}_{\text{para}}$ & \bf98.0 & 92.9 & \bf95.5 & 81.8 & 80.8 & 81.3 & \bf92.9 & 85.9 & \bf89.4 & \bf88.7 \\
  $\hat{O} \rightarrow \text{LED}_{\text{para}}$ & 90.9$^{\ast}$ & \bf96.0 & 93.4 & 80.8 & \bf82.8 & \bf81.8 & 83.8$^{\ast}$ & 90.9 & 87.4 & 87.5 \\
  $\hat{O} \rightarrow \text{ConBART}$ & 89.9$^{\ast\ast}$ & \bf96.0 & 92.9 & 81.8 & 79.8 & 80.8 & 86.9 & \bf91.9 & \bf89.4 & 87.7 \\
  \bottomrule
  \end{tabular}
  \caption{\textbf{Human evaluation results for selected simplification systems.} The \textit{minor} group includes those examples with a reading-level transition of 2 levels (e.g. 0-2, 1-3, etc.), whereas the \textit{major} class includes those of 3-4 levels. Each of these groups make up half of the entire set. Ratings significantly different from the highest score in each column are denoted with $\ast$ ($p < 0.05$) and $\ast\ast$ ($p < 0.01$). Significance was determined with two proportion $Z$-tests.}
  \label{tab:hum_eval}
\end{table*}

\section{Evaluation}
\subsection{Automatic Evaluation}
Text simplification is often evaluated on the basis of 3 criteria: adequacy (or meaning preservation), fluency, and simplicity. For automatic evaluation, we use BARTScore~\citep{bartscore_neurips21} and SMART~\citep{smarteval} as analogs for both adequacy and fluency. Both are reference-based metrics that have previously been used for document simplification as well as other text generation tasks.

For assessing simplicity, we use both the Flesch-Kincaid grade level (FKGL) and SARI~\citep{xu-etal-2016-optimizing}. FKGL is a document-level metric of text readability that has the highest correlation with human judgements~\citep{questeval_simp}, while SARI is a simplification metric that has become a staple in the sentence-level simplification literature. We use EASSE~\citep{alva-manchego-etal-2019-easse} to calculate both of these.

At test time we generate sequences using beam search with a beam size of 5 and a maximum length of 1024 tokens.

\subsection{Human Evaluation}
Historically, automatic evaluation of long-form text generation has been very difficult to perform \cite{howcroft-etal-2020-twenty,thomson-reiter-2020-gold}. As such, we conduct a human-evaluation of proposed systems to more accurately gauge performance. 

As full documents are very long and difficult to compare, we conduct evaluations at the paragraph-level. For each comparison, a complex paragraph is shown next to an extract from a generated simplification corresponding to that paragraph. Evaluators are then asked to judge whether the generated text (i) is fluent (\textbf{fluency}); (ii) preserves the core meaning of the input (\textbf{adequacy}); and (iii) is simpler to read/understand (\textbf{simplicity}).

Using the test set, we randomly sample 33 complex paragraphs from each non-adjacent reading-level transition pairing, for a total of 198 paragraphs. We take the references and outputs from 4 high performing systems ($\text{PG}_{\text{Dyn}}$, $\text{LED}_{\text{para}}$, $\hat{O} \rightarrow \text{LED}_{\text{para}}$, $\hat{O} \rightarrow \text{ConBART}$) for each (990 outputs in total) and have an annotator rate them on each of the 3 criteria. Because we use a large pool of annotators we impose a binary answering scheme (yes/no) in order to avoid the inter-annotator subjectivity that is inherent when using a Likert scale. The proportion of positive results is used as the final score for a given system.

Further details of the human evaluation are given in Appendix~\ref{app:human_eval}.

\section{Results and Discussion}
Results are shown in Table~\ref{tab:simp_eval}. We also report results for other commonly used metrics in Appendix~\ref{app:extra_results}.

\paragraph{Context Awareness Matters.} Considering all metrics, we find that text-only models that take as input either a sentence ($\text{BART}_{\text{sent}}$) or a whole document ($\text{BART}_{\text{doc}}, \text{LED}_{\text{doc}}$) underperform models whose input is more local to the input sentence, either because they work at the paragraph level ($\text{LED}_{\text{para}}$) or because they take both the complex sentence and its local document context as input (ConBART). In other words, models that have access to a \textit{local document context} ($\text{LED}_{\text{para}}$, ConBART) perform best overall.

\paragraph{LED vs BART.} 
LED models ($\text{LED}_{\text{doc/para}}$) outperform their standard counterpart ($\text{BART}_{\text{doc/para}}$) showing that modified self-attention is not only more efficient but also more precise than standard self-attention in the case of long input.
%This is likely because the full attention mechanism is unnecessary for the task, leading to blurry attention. Because of this change, LED models also perform inference much faster than $\text{BART}_{\text{doc/para}}$ (see Table~\ref{tab:model_speed}).

%% \paragraph{Sentences vs. Paragraphs}
%% When using the BART architecture, it seems paragraph-level models are fairly competitive with sentence-level ones ($\text{BART}_{\text{sent}}$ vs $\text{BART}_{\text{para}}$; $\text{PG}_{\text{Dyn}}$ vs $\hat{O} \rightarrow \text{BART}_{\text{para}}$), with some obvious differences. Paragraph-level models appear to do better on recall-based metrics but less so on precision.

%% However, this tradeoff appears to be eliminated when using $\text{LED}_{\text{para}}$, which outperforms sentence-level equivalents across the board. This suggests that having access to a wider window of token-level context at generation time allows for higher quality simplifications, provided that the attention is focused enough.

%% \paragraph{ConBART}
%% ConBART yields very good scores across the board, outperforming most other systems on every metric except FKGL. This shows that having access to the same high-level document structure information as previously proposed planners also directly benefits generative models. Further, when guided by a plan, ConBART sees even further improvements to performance.

\paragraph{The Utility of Planning.} Plan guided models (4th horizontal block in Table~\ref{tab:simp_eval}) outperform their standard couterpart on all metrics, showing that a predicted plan has a positive impact on simplification. This is further supported by the fact that models guided by an oracle plan (5th block) provide even greater performance.

\paragraph{Comparison with the State-of-the-Art ($\text{PG}_{\text{Dyn}}$).} $O \rightarrow \text{ConBART}$ is similar to $\text{PG}_{\text{Dyn}}$ in that, in both cases, a document is simplified by iterating over its sentences and prediction is guided by the local context of the sentence to be simplified. A key difference is that in $\text{PG}_{\text{Dyn}}$, this context is exclusively used to predict a simplification operation, while in $O \rightarrow \text{ConBART}$ it is additionally used to condition the generation of the simplified sentences. We find that adding this extra control results in significantly better scores compared to the state-of-the-art $\text{PG}_{\text{Dyn}}$ model. This illustrates that document context has utility for both planning (predicting the correct simplification operation) and realisation (simplifying a given sentence). While $\hat{O} \rightarrow \text{LED}_{\text{para}}$ achieves the best overall results of any system, it is slightly outperformed by $O \rightarrow \text{ConBART}$ when oracle plans are used, suggesting that an improved planner would provide better simplifications when used by $\hat{O} \rightarrow \text{ConBART}$ over $\hat{O} \rightarrow \text{LED}_{\text{para}}$.

%Guiding models that themselves also have access to some form of inter-sentence document context (ConBART, $\text{LED}_{\text{para}}$) 

\paragraph{Human Evaluation} Results from the human evaluation are shown in Table~\ref{tab:hum_eval}. To better identify where each model excels, we report separate scores for test paragraph pairs with minor (reading-level transition of 2) and major (>2) degrees of simplification, as well as total average scores. 

\begin{figure}[!htbp]
  \centering
  \begin{subfigure}[b]{0.48\textwidth}
    \centering
    \includegraphics[width=\textwidth]{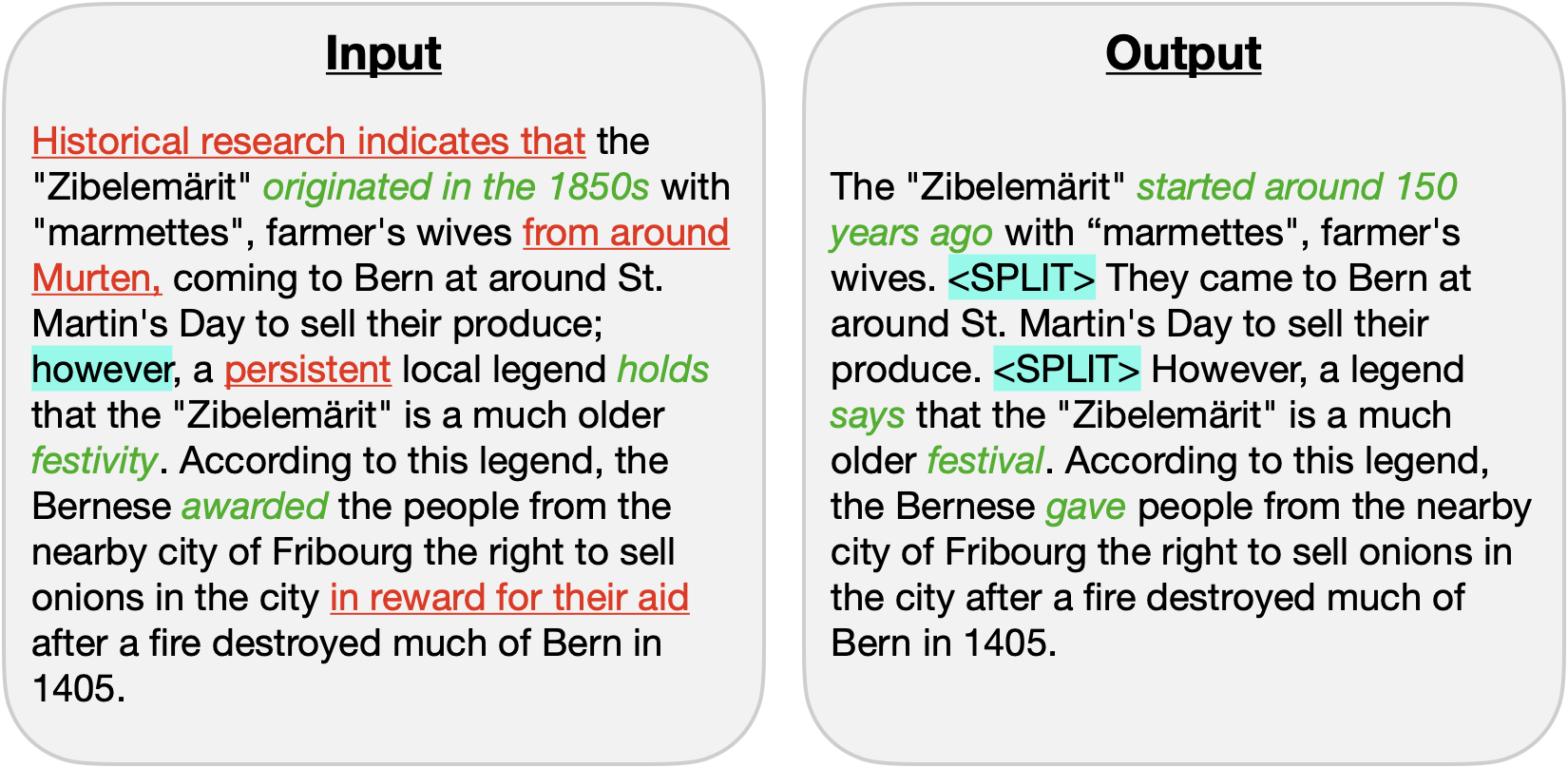}
    \caption{a good result}
    \label{fig:para_outputs_a}
  \end{subfigure}
  \begin{subfigure}[b]{0.48\textwidth}
    \centering
    \includegraphics[width=\textwidth]{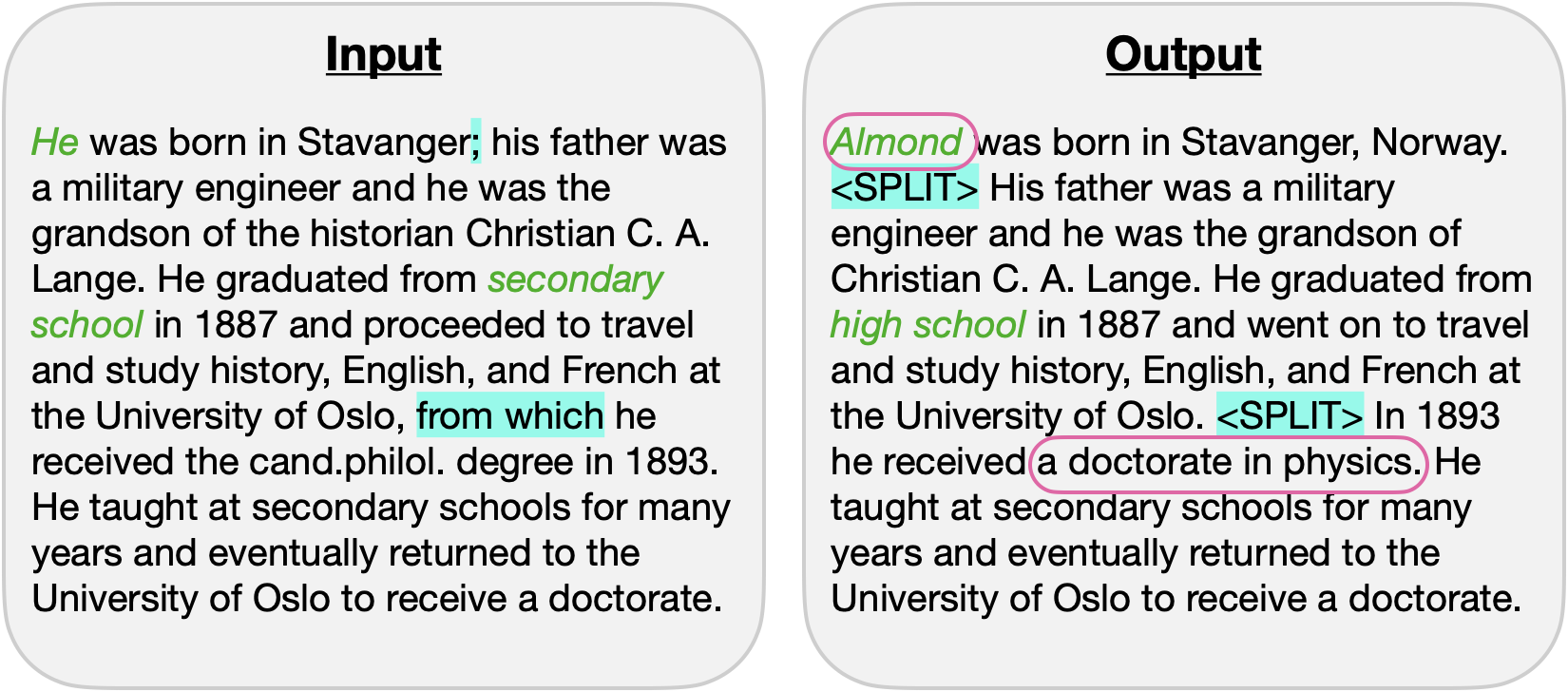}
    \caption{a poor result}
    \label{fig:para_outputs_b}
  \end{subfigure}
  \caption{\textbf{Example WikiLarge simplification output extracts from $\hat{O} \rightarrow \text{LED}_{\text{para}}$} (a target reading-level of 3 was used in each case). Note that these are small extracts from larger documents shown in Appendix~\ref{app:example_outputs}. Deletions are \textcolor{red}{\underline{underlined and in red}}; rephrasings are \textcolor{Green}{\it italicised and in green}; splitting points are \hlcyan{highlighted in cyan}; and factual errors are circled.}
  \label{fig:para_outputs}
\end{figure}

On fluency, all of the systems achieve very high ratings, which is unsurprising given the recognised ability of large language models (LLMs) to produce highly fluent texts. For adequacy, $\hat{O} \rightarrow \text{LED}_{\text{para}}$ achieves the highest overral score, closely followed by $\text{LED}_{\text{para}}$ and $\hat{O} \rightarrow \text{ConBART}$. In terms of simplicity, $\text{LED}_{\text{para}}$ and $\hat{O} \rightarrow \text{ConBART}$ equally achieve the highest score. Across all criteria, $\text{LED}_{\text{para}}$ achieves the highest average ratings, although very few scores are significantly better than other systems.

When considering performance differences between the minor and major simplification groups, we observe some clear trends. Systems that are not guided by a high-level document plan or do not have access to some contextual information during generation ($\text{PG}_{\text{Dyn}}$ and $\text{LED}_{\text{para}}$) perform notably worse on examples requiring major simplification than they do on minor cases. Conversely, the models with both of these features appear to either perform equally as well or even excel on major cases. This suggests potential conservativity in the simplifications performed by $\text{PG}_{\text{Dyn}}$ and $\text{LED}_{\text{para}}$.

Another interesting observation is the relatively low ratings given to the references compared to the system outputs. In particular, they receive a much lower adequacy score than any other system on major cases. This could perhaps be a result of the systems generating outputs that bear more of a resemblance to the inputs than those written by humans (see faithfulness BARTScores in Appendix~\ref{app:extra_results}). For instance, human editors might have been able to confidently delete more content, or refer to some of the information in different paragraphs which the evaluators were not privy to. Despite this, the references still receive fluency ratings competitive with the other systems.

\paragraph{Example Simplifications}
Figure~\ref{fig:para_outputs} shows some example simplification outputs from $\hat{O} \rightarrow \text{LED}_{\text{para}}$. These are paragraph-level extracts from larger document outputs, which are provided in Appendix~\ref{app:example_outputs}. Due to licensing constraints imposed by Newsela, we use out-of-domain documents from the WikiLarge dataset \citep{zhang-lapata-2017-sentence} in these examples.

\section{Model Efficiency}
There are various others factors to consider when comparing systems, beyond their raw performance. For instance, the size of the model(s) and how much time/resources are required for each to perform inference are important practical considerations that must be made when selecting a model for real-world use. As such, we compare each system based on the time taken to simplify the test set and their total parameter counts. Table~\ref{tab:model_speed} shows these results.

In our case, any system that uses a plan requires a second model, approximately doubling the number of parameters that must be loaded. These pipeline setups also naturally add to overall inference time. Further, both plan pipelines and ConBART make use of dynamic context, which imposes an autoregressive bottleneck on the simplification of individual documents. 

Because of the linearly scaling attention mechanism, Longformer-based models are the fastest of proposed systems. Because of this and its overall high performance, we recommend $\text{LED}_{\text{para}}$ in situations where time or computing resources are at all limited. Alternatively, $\hat{O} \rightarrow \text{ConBART}$ offers a good compromise that provides the high performance of a plan-guided system while mitigating further increases to inference time. This is because it uses the same autoregressive protocol as the planner and can therefore share the generated context representations.

All inference processes were run on a single Nvidia A40 GPU, using a batch size of 16, 32 CPU workers for data loading, and a beam size of 5 for generation. Appendix~\ref{app:dyncon_alg} provides details on the specific algorithm used to handle dynamic context generation for appropriate models.

\begin{table}[!htbp]
  \centering
  \small
  \begin{tabular}{lcc}
    \toprule
    & \bf Inference Time $\downarrow$ & \bf{\# Params $\downarrow$} \\
    \midrule
    $\text{BART}_{\text{doc}}$ & 182.6 & 140 \\
    $\text{BART}_{\text{sent}}$ & 54.0 & 140 \\
    $\text{BART}_{\text{para}}$ & 68.9 & 140 \\
    $\text{LED}_{\text{doc}}$ & 49.1 & 162 \\
    $\text{LED}_{\text{para}}$ & 45.9 & 162 \\
    $\text{ConBART}$ & 74.7 & 156 \\
    \midrule
    $\text{PG}_{\text{Dyn}}$ (\citeyear{cripwell-etal-2023-document}) & 76.6 & 154+140 \\
    $\hat{O} \rightarrow \text{BART}_{\text{para}}$ & 119.1 & 154+140 \\
    $\hat{O} \rightarrow \text{LED}_{\text{para}}$ & 103.3 & 154+162 \\
    $\hat{O} \rightarrow \text{ConBART}$ & 82.7 & 154+156 \\
    \bottomrule
  \end{tabular}
  \caption{\textbf{Model efficiency statistics.} All times are in milliseconds and model parameters are in millions. Inference times are calculated on the test set and normalised by the total number of sentences (i.e. \# ms per sentence).}
  \label{tab:model_speed}
\end{table}

\section{Conclusion}
We develop a range of document simplification models that are able to use different combinations of text, context, and simplification plans as input, with several models outperforming the previous state-of-the-art both on  automatic metrics and according to human judgements. Our results show that a high-level representation of the document can be useful for low-level surface realisation as well as global planning. Further, simplification models with access to local document context, either by working at the paragraph level or handling an additional input representation, lead to better meaning preservation than those that operate on individual sentences. We conclude by evaluating the model efficiency of each system and making recommendations for their selection under different circumstances.

\section{Limitations}

\paragraph{Newsela Dataset}
One limitation to this study is our use of the Newsela dataset. Because this requires a license to access, researchers cannot fully reproduce our work without first obtaining permission from Newsela Inc. Unfortunately there is currently no other large dataset offering high quality aligned documents for simplification under an open source license. The only other datasets so far used for document-level simplification are based on WikiLarge, which has very poor and inconsistent alignments at the document-level \cite{xu-etal-2015-problems,sun-etal-2021-document,cripwell-etal-2023-document}.

\paragraph{Paragraph-Level Human Evaluation}
In order to reduce complexity, our human evaluation was performed on paragraphs rather than full documents. As a result, there is a potential limit to the accuracy of human judgements when certain discourse phenomena are present. For example, important information may be excluded from a specific output paragraph (therefore prompting a low adequacy rating), but this could actually be present in a different part of the true simplified document.

\paragraph{Monolinguality}
This study focused entirely on simplification for English-language documents. Reproducing the proposed systems for use on other languages would require dedicated datasets of similar scale, along with sentence/paragraph alignments and operation labels (which likely do not currently exist). Further, the nature of simplification in other languages may differ quite a lot from English with respect to the types of operations that are performed, potentially reducing the suitability of the proposed framework.

\paragraph{Generalised Target Audience}
We approach this study with our definition of "simplification" being based on that of a generalised audience, following the standard set out by the assigned reading-levels of the Newsela dataset. Existing works often outline the intent for their systems to be used to simultaneously assist a wide array of different target users, such as those with cognitive impairments, non-native speakers, and children~\citep{maddela-etal-2021-controllable,garbacea-etal-2021-explainable,sun-etal-2021-document}. However, they rarely go into any detail about which simplification strategies work for each of these different groups or perform human evaluation with annotators from the same target demographics~\citep{gooding-2022-ethical}. As such, we acknowledge that using our systems for a specific demographic might prove insufficient to enable their consumption of media without first making further revisions to support their precise needs.

%% \section*{Acknowledgments}
%% We thank the anonymous reviewers for their feedback. We gratefully acknowledge the support of the French National Research Agency (Gardent; award ANR-20-CHIA-0003, XNLG "Multi-lingual, Multi-Source Text Generation") and of Facebook AI Research (FAIR) Paris.

%% Experiments presented in this paper were carried out using the Grid’5000 testbed, supported by a scientific interest group hosted by Inria and including CNRS, RENATER and several Universities as well as other organizations (see \url{https://www.grid5000.fr}).

\bibliography{anthology,custom}

\appendix

\section{Training Details}
\label{app:train_details}
For all simplification models, we used a learning rate of $2e^{-5}$, a batch size of 16, and a 0.1 dropout rate. All models were trained on a computing grid using $2 \times$ Nvidia A40 GPUs (45GB memory) until convergence or a maximum of 48 hours.

For ConBART and planning pipelines we use the same settings as \citet{cripwell-etal-2023-document} for construction of the high-level document context. Specifically, this includes a fixed context window radius of size 13 and use of a dynamic context mechanism.

\section{Multi-Task Systems}
\label{app:mtl}
We also experimented with models that are explicitly trained to perform both the planning and simplification tasks using the same network. As high-level plans appear to improve the performance of simplification models, we hypothesise that learning both tasks in tandem could benefit overall performance. The motivation for this approach is to potentially produce a model that is capable of yielding similar or better simplification performance to the pipeline systems but with a more efficient single-model setup.

Specifically, these models were trained to generate the simplified text prefixed by a predicted plan in the form of operation-specific tokens. This was tested with both ConBART (\textbf{$\text{ConBART}_{\text{prefix}}$}) and a document-level Longformer (\textbf{$\text{LED}_{\text{prefix}}$}). In the case of the Longformer we also test a variant that generates the plan tokens as sentence separators (\textbf{$\text{LED}_{\text{sep}}$}). Results are shown in Table~\ref{tab:mtl_simp_eval}.

Unfortunately, from our experiments none of these seemed to result in performance exceeding those of simplification-only models. Improvement could perhaps be reached given the correct tuning of hyperparameters and loss weightings, however we did not have the time or resources to pursue this further in this study.

\begin{table*}[!htbp]
  \centering
  \small
  \begin{tabular}{lcccc|ccc|cc|cc}
  \toprule
  \bf System & \multicolumn{4}{c}{\bf BARTScore} & \multicolumn{3}{c}{\bf SMART $\uparrow$} & \bf FKGL $\downarrow$ & \bf SARI $\uparrow$  & \multicolumn{2}{c}{\bf Length} \\
  \midrule
    & Faith. & P $\uparrow$ & R $\uparrow$ & F1 $\uparrow$ & P & R & F1 &  &  & Tok. & Sent. \\
    & ($s \rightarrow h$) & ($r \rightarrow h$) & ($h \rightarrow r$) & & & & & & & &  \\
  \midrule
  Input & -0.93 & -2.47 & -1.99 & -2.23 & 63.2 & 62.7 & 62.8 & 8.44 & 20.52  & 866.9 & 38.6 \\
  Reference & -1.99 & -0.93 & -0.93 & -0.93 & 100 & 100 & 100 & 4.93 & 99.99 & 671.5 & 42.6 \\
  \midrule
  $\text{LED}_{\text{prefix}}$ & -1.83 & -1.72 & -2.00 & -1.86 & 73.5 & 67.6 & 69.8 & 4.97 & 63.14 & 604.0 & 38.0 \\
  $\text{LED}_{\text{sep}}$ & -1.82 & -1.80 & -1.88 & -1.84 & 72.6 & 70.5 & 71.1 & 5.06 & 62.64 & 640.4 & 40.2 \\
  $\text{ConBART}_{\text{prefix}}$ & -1.96 & -1.62 & -1.60 & -1.61 & 80.4 & 79.6 & 79.9 & 4.90 & 74.31 & 643.6 & 41.5 \\
  \bottomrule
  \end{tabular}
  \caption{Results of multi-task systems on the Newsela-auto test set.}
  \label{tab:mtl_simp_eval}
\end{table*}

\section{Human Evaluation Details}
\label{app:human_eval}
The Newsela-auto paragraph alignments were used to identify valid references for each test paragraph. In order to align correct extracts from generated system outputs we took different steps depending on the system. For paragraph-level models (those using $\text{LED}_{\text{para}}$), we simply use the full simplification output for each source paragraph. For sentence-level models (ConBART, $\text{PG}_{\text{Dyn}}$), we first used the alignments to identify which paragraph the source sentence belongs to, then concatenated their simplification results.

\begin{figure*}[!htbp]
  \centering
  \includegraphics[width=0.95\textwidth]{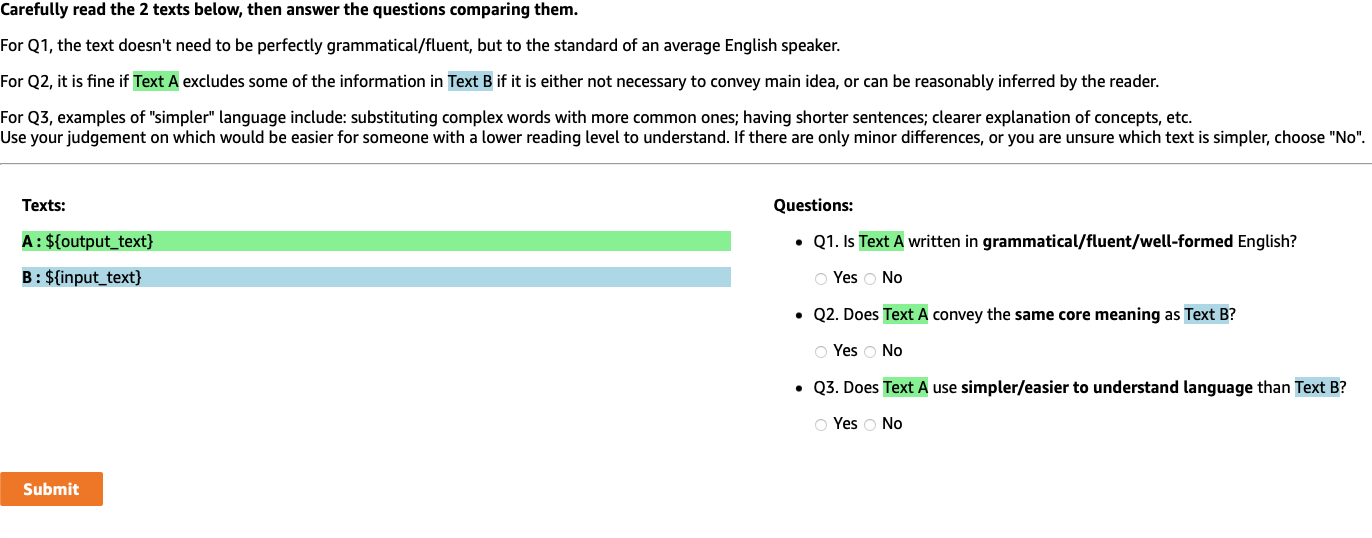}
  \caption{Submission form used in human evaluation.}
  \label{fig:human_eval_form}
\end{figure*}

Human judgements were crowdsourced on the MTurk platform. We sourced workers from English speaking countries (AU, CA, GB, IE, NZ, US) and paid them \$0.2 USD for each individual evaluation. We ran an initial test ourselves and timed how many evaluations could be completed within an hour. According to this, subjects should earn approximately \$18 USD per hour (which is above the minimum wage in all of these countries). The form and instructions presented to human evaluators is shown in Figure~\ref{fig:human_eval_form}.

\section{Additional Evaluation Results}
\label{app:extra_results}

In Table~\ref{tab:simp_eval_extra} we provide additional results for popular automatic evaluation metrics that were not included in the main text. Specifially, we include BLEU~\citep{papineni-etal-2002-bleu}, and full operation-specific scores for SARI. In general, the results are similar to those in Table~\ref{tab:simp_eval}, with $\hat{O} \rightarrow \text{LED}_{\text{para}}$ and $\hat{O} \rightarrow \text{ConBART}$ achieving the best results. 

Faithfulness BARTScore is included for clarity rather than being a direct estimation of output quality. It shows how semantically similar system outputs are to their inputs, roughly equating to a measurement of conservativity.

\begin{table*}[!htbp]
  \centering
  \small
  \begin{tabular}{lclllll}
  \toprule
  \bf System & \bf BARTScore & \bf BLEU $\uparrow$ & \bf SARI $\uparrow$ & add & keep & delete \\
  \midrule
    & Faith. ($s \rightarrow h$) \\
  \midrule
  Input & -0.93 & 46.2 & 20.5 & 0.0 & 61.6 & 0.0 \\
  Reference & -1.99 & 100 & 100 & 100 & 100 & 100 \\
  \midrule
  $\text{BART}_{\text{doc}}$ & -2.48 & 31.1 & 47.1 & 20.4 & 55.4 & 65.4 \\
  $\text{BART}_{\text{sent}}$ & -1.86 & 70.7 & 73.0 & 55.9 & 83.7 & 79.5 \\
  $\text{BART}_{\text{para}}$ & -2.11 & 68.6 & 73.7 & 57.8 & 82.6 & 80.8 \\
  $\text{LED}_{\text{doc}}$ & -1.90 & 63.7 & 68.7 & 52.2 & 78.2 & 75.7 \\
  $\text{LED}_{\text{para}}$ & -1.86 & 74.5$^{\ast}$ & 76.9$^{\ast}$ & 64.3$^{\ast}$ & 85.0 & 81.5 \\
  $\text{ConBART}$ & -1.89 & 73.7 & 75.8 & 61.4 & 84.9 & 81.2 \\
  \midrule
  $\text{PG}_{\text{Dyn}}$ (\citeyear{cripwell-etal-2023-document}) & -1.91 & 72.4 & 75.0 & 58.9 & 84.8 & 81.4 \\
  $\hat{O} \rightarrow \text{ConBART}$ & -1.92 & 76.0$^{\ast}$ & 78.3$^{\ast}$ & 64.6$^{\ast}$ & 86.8$^{\ast}$ & \bf83.4 \\
  $\hat{O} \rightarrow \text{BART}_{\text{para}}$ & -2.05 & 71.3 & 74.9 & 58.5 & 84.7 & 81.4 \\
  $\hat{O} \rightarrow \text{LED}_{\text{para}}$ & -1.87 & \bf76.8$^{\ast}$ & \bf78.5$^{\ast}$ & \bf65.1$^{\ast}$ & \bf87.3$^{\ast}$ & 83.0 \\
  \midrule
  $\text{PG}_{\text{Oracle}}$ (\citeyear{cripwell-etal-2023-document}) & -1.93 & 78.9$^{\ast}$ & 80.7$^{\ast}$ & 65.2$^{\ast}$ & 89.9$^{\ast}$ & 87.1$^{\ast}$ \\
  $O \rightarrow \text{ConBART}$ & -1.93 & \bf82.6$^{\ast}$ & \bf83.8$^{\ast}$ & \bf70.8$^{\ast}$ & \bf91.7$^{\ast}$ & \bf88.7$^{\ast}$ \\
  $O \rightarrow \text{BART}_{\text{para}}$ & -2.09 & 76.1$^{\ast}$ & 79.7$^{\ast}$ & 64.1$^{\ast}$ & 88.7$^{\ast}$ & 86.3$^{\ast}$ \\
  $O \rightarrow \text{LED}_{\text{para}}$ & -1.90 & 81.4$^{\ast}$ & 82.3$^{\ast}$ & 69.6$^{\ast}$ & 90.6$^{\ast}$ & 86.7$^{\ast}$ \\
  \bottomrule
  \end{tabular}
  \caption{Extra automatic evaluation results on Newsela-auto. For BARTScore, $s$ is the source text and $h$ is the hypothesis. Scores significantly higher than $\text{PG}_{\text{Dyn}}$ are denoted with $\ast$ ($p < 0.005$). Significance was determined with Student's $t$-tests.}
  \label{tab:simp_eval_extra}
\end{table*}

\section{Example Simplifications}
\label{app:example_outputs}
\begin{figure*}[!htbp]
  \centering
  \includegraphics[width=0.85\textwidth]{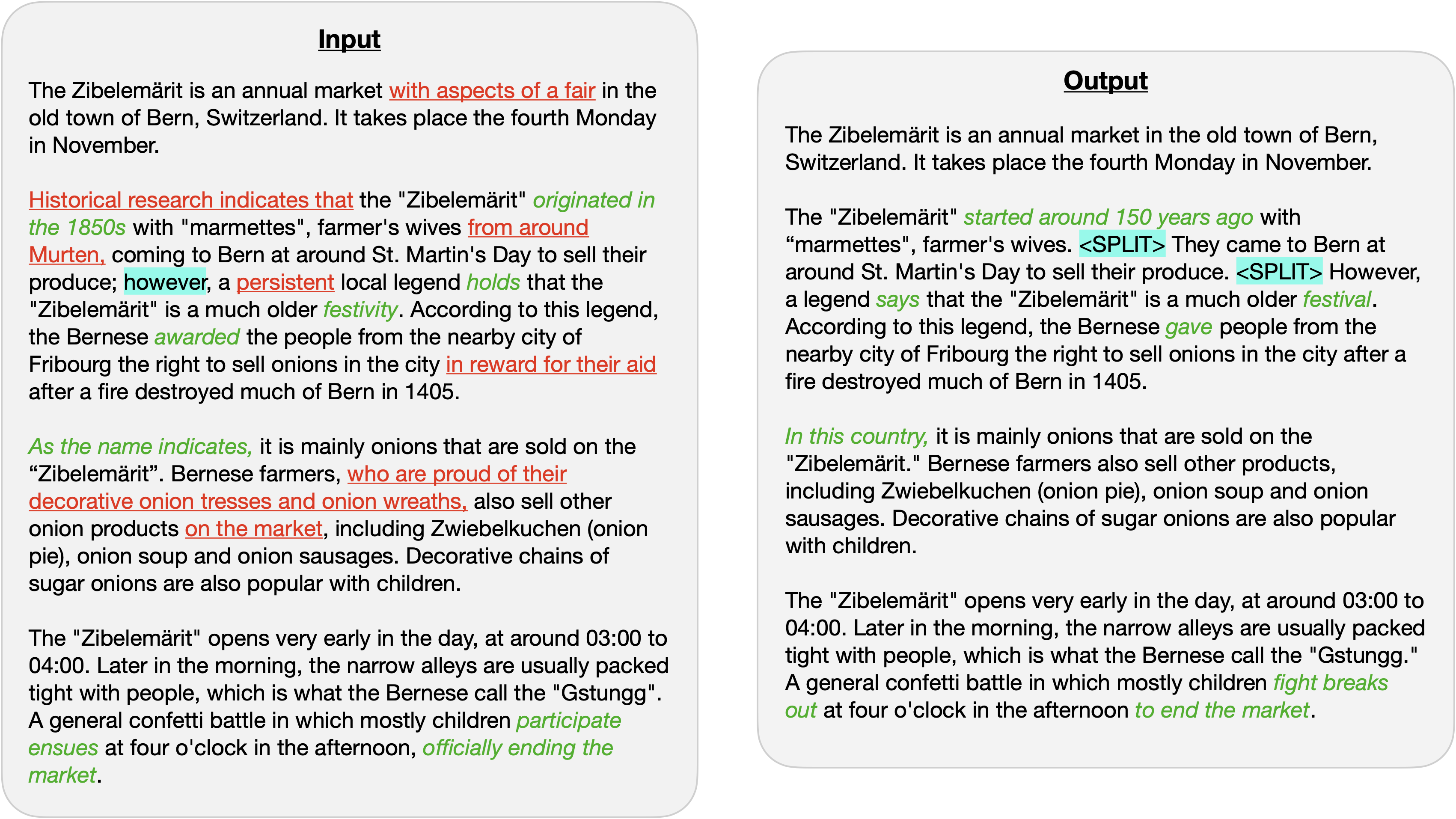}
  \includegraphics[width=0.85\textwidth]{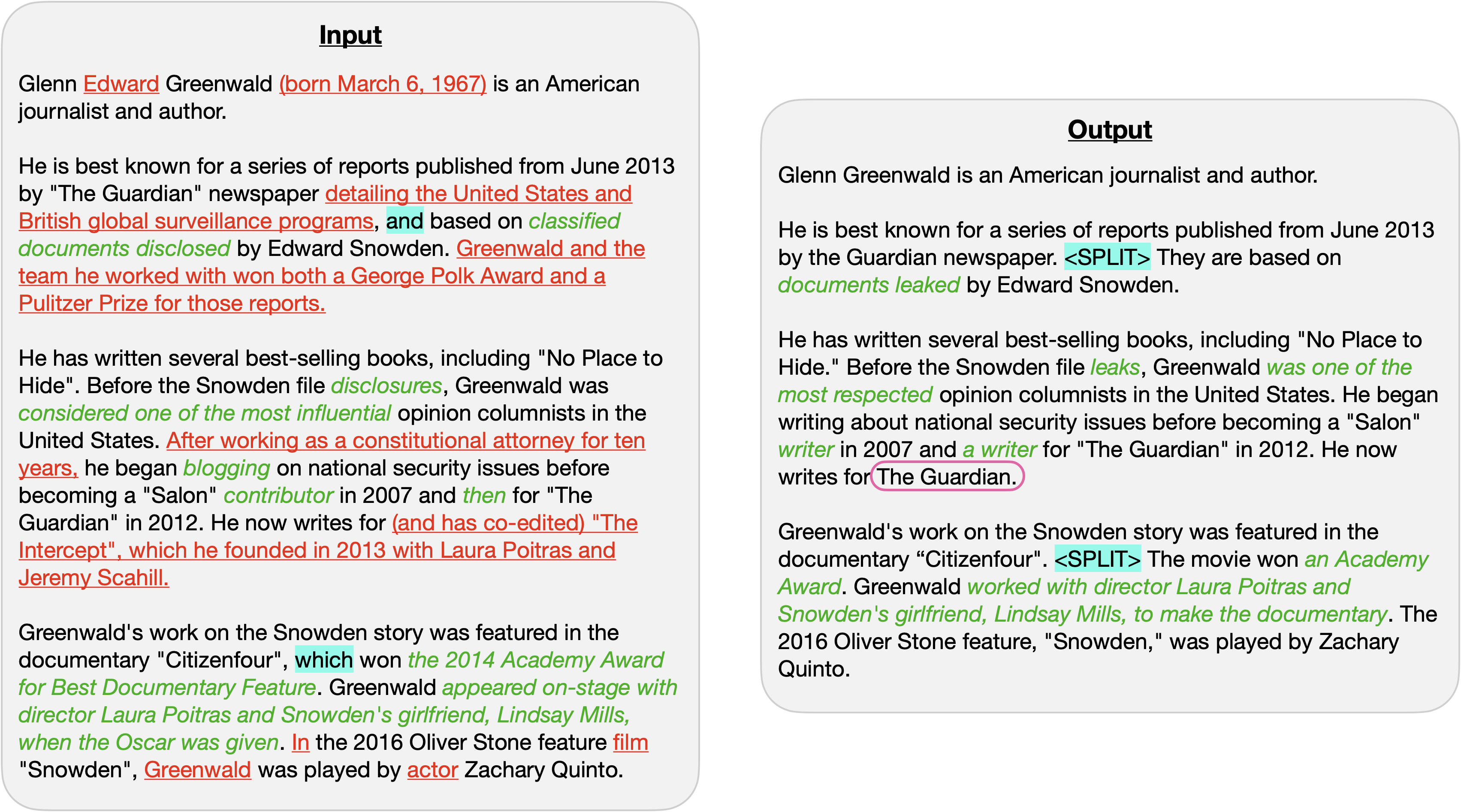}
  \includegraphics[width=0.85\textwidth]{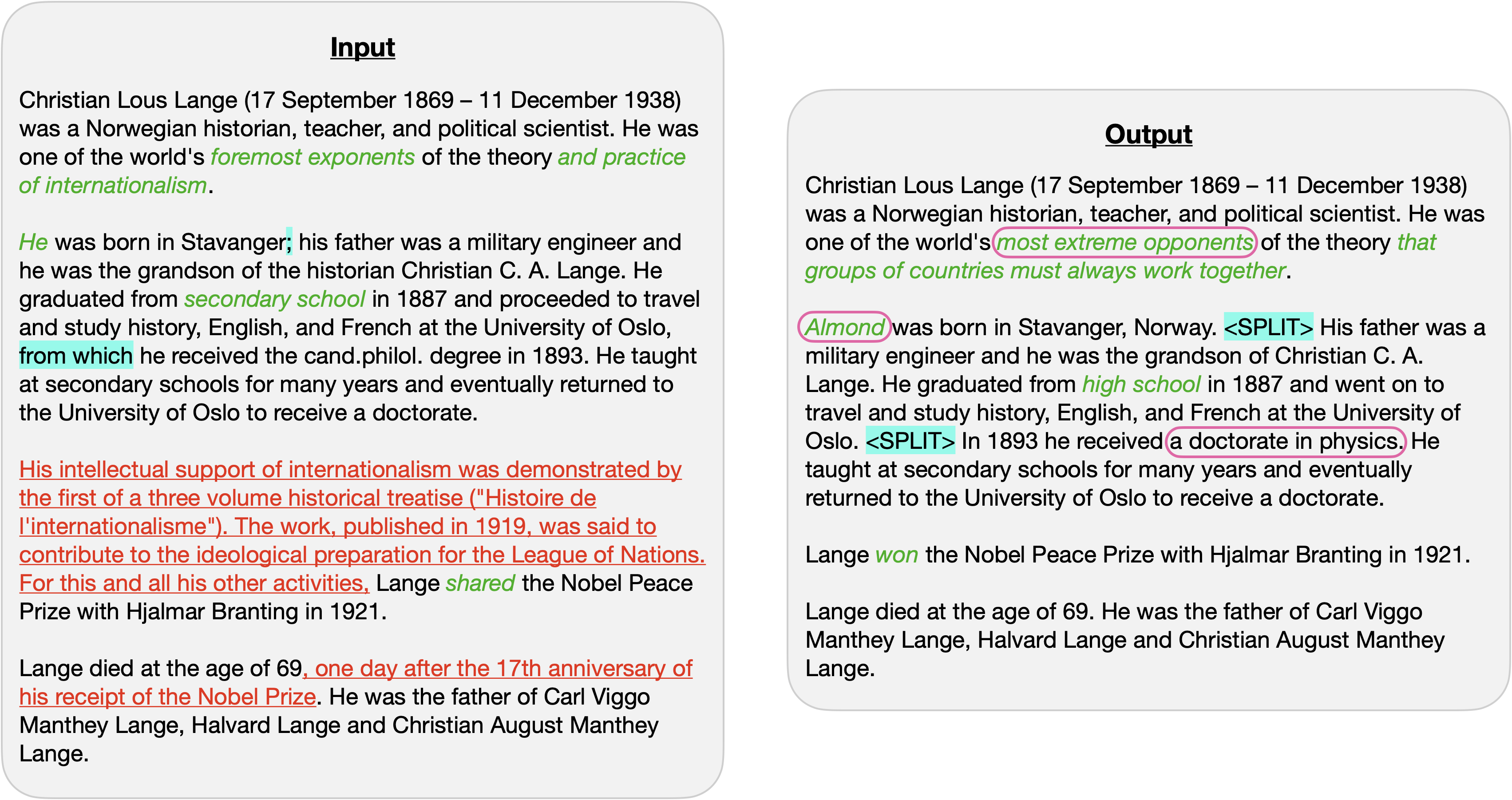}
  \caption{Example simplification outputs for $\hat{O} \rightarrow \text{LED}_{\text{para}}$, illustrating both strong and poor performances (a target reading-level of 3 was used for all examples). Input documents are taken from WikiLarge due to licensing constraints around sharing Newsela content. Deletions are \textcolor{red}{\underline{underlined and in red}}; rephrasings are \textcolor{Green}{\it italicised and in green}; splitting points are \hlcyan{highlighted in cyan}; and factual errors are circled.}
  \label{fig:doc_outputs}
\end{figure*}

Figure~\ref{fig:doc_outputs} shows several example simplifications by the $\hat{O} \rightarrow \text{LED}_{\text{para}}$ system on full documents. Due to licensing constraints imposed by Newsela, we use out-of-domain documents from the WikiLarge dataset here. As these are Wikipedia articles they are quite different in tone than the Newsela articles as well as being much shorter in length. Regardless, we still believe this provides clarity on the types of editing performed by the model.

\section{Dynamic Context Algorithm}
\label{app:dyncon_alg}
Algorithm~\ref{alg:dyngen} shows the process used to handle dynamic context generation for appropriate models. As each document needs to be simplified autoregressively at the sentence level, we construct batches of sentences with the same index from different documents in order to speed up processing. Note that this could potentially be further optimised (e.g. via parallelism) and merely serves as a reasonable baseline algorithm.

\begin{algorithm*}
  \caption{Generation strategy for systems using a dynamic context mechanism. Inference is performed autoregressively in batches containing 1 sentence per document. At the end of each time step, the simplified sentences are encoded for use within the context of the next step. This naturally extends to the paragraph-level case by replacing sentences with paragraphs.}
  \label{alg:dyngen}
  \begin{algorithmic}[1]
    \Procedure{DynamicGeneration}{$\textit{test\_set}$}
      \State $g \gets \textit{load\_planner}()$
      \State $h \gets \textit{load\_simplifier}()$
      \State $\textit{max\_idx} \gets \max\limits_{C \in \textit{test\_set}}{|C|}$
      \For{$i \gets 1$ to $\textit{max\_idx}$}
        \State $\textit{sents} \gets \{c_i \mid C \in \textit{test\_set}\}$ \Comment{$i$th sentence from each document}
        \State $\textit{context} \gets \textit{load\_context}(\textit{sents})$
        \If{pipeline system}
          \State $\textit{plans} \gets g(\textit{sents},\textit{context})$
          \State $\textit{sents} \gets \textit{plans} + \textit{sents}$ \Comment{Prepend plans to texts}
        \EndIf
        \State $\textit{preds} \gets h(\textit{sents,context})$
        \State $\textit{context} \gets \textit{update\_context}(\textit{preds})$
      \EndFor
    \EndProcedure
  \end{algorithmic}
\end{algorithm*}

\end{document}